\documentclass[sigconf]{acmart}

\AtBeginDocument{%
  }

\setcopyright{acmlicensed}
\copyrightyear{2026}
\acmYear{2026}
\acmDOI{XXXXXXX.XXXXXXX}
\acmConference[RecSys '26]{Proceedings of the 20th ACM Conference on Recommender Systems}{September 2026}{TBD}
\acmISBN{978-1-4503-XXXX-X/2026/09}

\usepackage{style_overwrite_preprint}

\newcommand{\showlocations}{}
\ifdefined\showlocations
  \newcommand{\authorlocation}[2]{\city{#1}\country{#2}}
\else
  \newcommand{\authorlocation}[2]{}
\fi

\DeclareUrlCommand\code{\urlstyle{tt}}

\usepackage{listings}
\usepackage{multirow}
\begin{document}

\title{WorkRB:  A Community-Driven Evaluation Framework for AI in the Work Domain}

\author{Matthias De Lange}
\authornote{Correspondence to \url{workrb@techwolf.ai}.}
\affiliation{\institution{TechWolf}\authorlocation{Ghent}{Belgium}}

\author{Warre Veys}
\affiliation{\institution{TechWolf}\authorlocation{Ghent}{Belgium}}

\author{Federico Retyk}
\affiliation{\institution{Avature}\authorlocation{Barcelona}{Spain}}

\author{Daniel Deniz}
\affiliation{\institution{Avature}\authorlocation{Barcelona}{Spain}}

\author{Warren Jouanneau}
\affiliation{\institution{Malt}\authorlocation{Paris}{France}}

\author{Mike Zhang}
\affiliation{\institution{University of Copenhagen}\authorlocation{Copenhagen}{Denmark}}

\author{Aleksander Bielinski}
\affiliation{\institution{Edinburgh Napier University}\authorlocation{Edinburgh}{UK}}

\author{Emma Jouffroy}
\affiliation{\institution{Malt}\authorlocation{Paris}{France}}

\author{Nicole Clobes}
\affiliation{\institution{WAPES}\authorlocation{Brussels}{Belgium}}%

\author{Nina Baranowska}
\affiliation{\institution{Leiden University}\authorlocation{Leiden}{Netherlands}}

\author{David Graus}
\affiliation{\institution{University of Amsterdam}\authorlocation{Amsterdam}{Netherlands}}

\author{Marc Palyart}
\affiliation{\institution{Malt}\authorlocation{Paris}{France}}

\author{Rabih Zbib}
\affiliation{\institution{Avature}\authorlocation{Barcelona}{Spain}}

\author{Dimitra Gkatzia}
\affiliation{\institution{Edinburgh Napier University}\authorlocation{Edinburgh}{UK}}

\author{Thomas Demeester}
\affiliation{\institution{Ghent University - imec}\authorlocation{Ghent}{Belgium}}

\author{Tijl De Bie}
\affiliation{\institution{AIDA-IDLab, Ghent University}\authorlocation{Ghent}{Belgium}}

\author{Toine Bogers}
\affiliation{\institution{IT University of Copenhagen}\authorlocation{Copenhagen}{Denmark}}

\author{Jens-Joris Decorte}
\affiliation{\institution{TechWolf}\authorlocation{Ghent}{Belgium}}

\author{Jeroen Van Hautte}
\affiliation{\institution{TechWolf}\authorlocation{Ghent}{Belgium}}

\renewcommand{\shortauthors}{De Lange et al.}

\begin{abstract}
Today's evolving labor markets rely increasingly on recommender systems for hiring, talent management, and workforce analytics, 
with natural language processing (NLP) capabilities at the core.
Yet, research in this area remains highly fragmented. Studies employ divergent ontologies (ESCO, O*NET, national taxonomies), heterogeneous task formulations, and diverse model families, making cross-study comparison and reproducibility exceedingly difficult.
General-purpose benchmarks lack coverage of work-specific tasks, and the inherent sensitivity of employment data further limits open evaluation. We present \textbf{WorkRB} (Work Research Benchmark), the first open-source, community-driven benchmark tailored to work-domain AI. WorkRB organizes 13~diverse tasks from 7~task groups as unified recommendation and NLP tasks, including job\,/\,skill recommendation, candidate recommendation, similar item recommendation, and skill extraction and normalization. WorkRB enables both monolingual and cross-lingual evaluation settings through dynamic loading of multilingual ontologies. Developed within a multi-stakeholder ecosystem of academia, industry, and public institutions, WorkRB has a modular design for seamless contributions and enables integration of proprietary tasks without disclosing sensitive data. WorkRB is available under the Apache~2.0
license at \url{https://github.com/techwolf-ai/WorkRB}.
\end{abstract}



\begin{teaserfigure}
    \centering
  \includegraphics[width=0.9\textwidth,
    trim=-8mm 0mm 1mm 2mm,
    clip
    ]{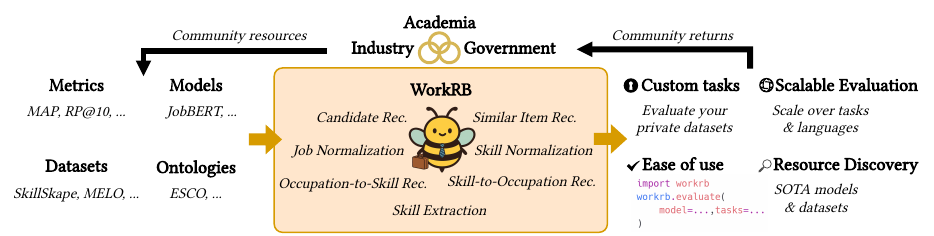}
  \caption{Overview of the community-driven WorkRB evaluation framework.}
  \Description{A high-level overview diagram of the WorkRB framework showing tasks, models, and multilingual evaluation.}
  \label{fig:teaser}
\end{teaserfigure}

\maketitle

\section{Introduction}

Work-domain AI capabilities, including job--skill recommendation, skill extraction, job normalization, and related tasks, power hiring platforms, talent management systems, and public employment services worldwide. However, research in this area remains highly fragmented: practitioners and researchers employ a wide variety of task formulations, languages, model families, and ontologies such as ESCO~\cite{esco}, O*NET~\cite{onet}, SkillsFuture~\cite{skillsfuture_skills_framework}, and numerous national taxonomies.
This fragmentation makes it exceedingly difficult to compare results across studies, reproduce findings, or build incrementally on prior work, slowing collective progress.
Each task typically requires its own bespoke evaluation setup, from dataset preparation to metric selection, prohibiting scaling across multiple work-domain tasks.
Moreover, work-domain data is inherently highly sensitive; career histories, compensation records, and employment data constitute personal data subject to strict privacy regulation, making it both undesirable and difficult to share datasets openly, which further limits reproducibility.
Meanwhile, multilingual occupational ontologies such as ESCO contain rich hierarchical and cross-lingual structures, that remain largely untapped for evaluation purposes.

While common NLP benchmarks such as MTEB~\cite{mteb}, BEIR~\cite{beir}, and SuperGLUE~\cite{superglue} focus on general-purpose capabilities, they are not aligned with recommendation tasks, evaluation strategies, and models specific to the work domain.

To this end, we introduce \textbf{WorkRB} (Work Research Benchmark), an open-source benchmark framework that unifies datasets, baseline models, and multilingual evaluation across work-related recommendation and NLP tasks.
Our contributions are as follows:
\begin{enumerate}
  \item \textbf{Unified benchmark.} WorkRB scales evaluation across 13 recommendation and NLP tasks and 7 task groups, evaluated in a unified fashion as ranking problems~\cite{workbench}.
  Through dynamic ontology loading, WorkRB currently supports up to 28~languages, matching ESCO's full language coverage, enabling both monolingual and cross-lingual tasks.
  \item \textbf{Extensible toolkit.} WorkRB is pip-installable and features a modular design with extendable base classes for tasks and models, automatic checkpointing, and hierarchical metric aggregation. 
  Open-source evaluation tasks can be complemented by proprietary, sensitive-data tasks for internal use.
  \item \textbf{Multi-stakeholder ecosystem.} As exemplified in Figure~\ref{fig:teaser}, WorkRB is a joint effort spanning academia, industry, and public institutions, bridging communities that each hold complementary pieces of the work-domain puzzle. Industrial and academic contributors provide real-world tasks, datasets, and domain-specialized models from published research; and government institutions supply extensive occupational ontologies (e.g., ESCO by the European Commission, O*NET by the US Department of Labor) that serve as the taxonomic backbone for many tasks. WorkRB provides a structured contribution model that incentivizes continued participation across all three communities.
    \item \textbf{Broader impact.} We discuss how open standardization supports legal compliance for sensitive work data, improves language representativeness, and enables standardized evaluation for public employment services.
\end{enumerate}

\section{The Work Research Benchmark}

WorkRB\footnote{WorkRB is under continuous development by community contributions. The version described in this paper is based on \emph{v0.5.1}.} is designed to accommodate multiple ontologies, task types, and languages within a comparable evaluation framework.
To avoid restricting evaluation to a single setup, WorkRB operates as a repository-style evaluation toolbox, providing a curated collection of datasets and baselines, with a focus on flexible configurations, rather than functioning as a constrained competition platform with a live leaderboard.

\subsection{Evaluation Tasks \& Ontologies}

WorkRB comprises 13 recommendation and NLP tasks organized into 7 task groups, addressing core retrieval and ranking scenarios in the work domain.
Following prior work~\cite{workbench}, all of the contributed work-domain tasks are formulated as ranking problems. 
However, WorkRB remains flexible in architecture, providing support for classification tasks, and can also be extended into other task formats.
Many tasks are grounded in an ontology, such as ESCO~\cite{esco}, for which WorkRB provides a flexible interface 
that automatically processes and caches the ontology for a given version and language, enabling efficient re-use across tasks.
Table~\ref{tab:tasks} provides a complete overview of all tasks with their label type, number of supported languages, and dataset size in terms of their number of queries and targets for recommendation.
We describe each task group below.
\begin{itemize}
  \item \textbf{Occupation-to-Skill Recommendation (SRec)} ranks ESCO skills for a given occupation, where each occupation maps to multiple relevant skills~\cite{esco,workbench}.
  \item \textbf{Skill-to-Occupation Recommendation (ORec)} ranks ESCO occupations for a given skill, where each skill maps to multiple relevant occupations~\cite{esco,workbench}.
  \item \textbf{Similar Item Recommendation (SIRec)} ranks skills~\cite{skillmatch1k} or occupation titles~\cite{jobtitle_sim_zbib2022Learning,jobtitle_sim_deniz2024Combined} by semantic relatedness, with single and multi-label targets respectively. 
  \item \textbf{Candidate Recommendation (CRec)} ranks candidate profiles in cross-lingual settings given a freelancer project description or keyword search query, with multiple relevant candidates per query~\cite{jouanneau2024skill,jouanneau2025efficient}. \looseness=-1
  \item \textbf{Job Normalization (JNorm)} maps free-text job titles or national taxonomy entries to standardized ESCO occupation entries, including cross-lingual entity linking 
  ~\cite{jobbert,retyk2024melo_mels}.
  \item \textbf{Skill Normalization (SNorm)} maps skill surface forms or national taxonomy entries to canonical ESCO skill entries, including cross-lingual entity linking~\cite{esco,workbench,retyk2024melo_mels}.
  \item \textbf{Skill Extraction (SExtr)} retrieves relevant skills from job descriptions, where queries are sentences ranked against a target skill taxonomy~\cite{skillextraction_decorte2022design,skillskape}.
\end{itemize}

\begin{table}[b]
  \caption{Overview of WorkRB tasks. Dataset sizes are shown for the English (en) number of queries $\times$ targets. For non-English datasets, we separately report the maximum number of queries and targets for cross-lingual (x), Bulgarian (bg), and Swedish (sv).
  ESCO-based target spaces vary by language and version. 
  }
  \label{tab:tasks}
  \Description{Table listing the 13 ranking tasks in WorkRB organized by task group, with their label type, number of queries and targets, and number of supported languages.}
    \resizebox{\columnwidth}{!}{%
  \begin{tabular}{lcll}
    \toprule
    \textbf{Dataset} & \textbf{Labels} & \textbf{Size (en)} & \textbf{Lang.} \\
    \midrule
    \textbf{Occupation-to-Skill Rec.} \\
     ESCO Occupation-to-Skill~\cite{esco,workbench} & multi & 3,039 (en) $\times$ 13,939 (en)& 28 \\
    \addlinespace
    \textbf{Skill-to-Occupation Rec.} \\
     ESCO Skill-to-Occupation~\cite{esco,workbench} & multi & 13,492 (en) $\times$ 3,039 (en)& 28 \\
     \addlinespace
    \textbf{Similar Item Rec.} \\
     Job Title Sim.~\cite{jobtitle_sim_zbib2022Learning,jobtitle_sim_deniz2024Combined} & multi & 105 (en) $\times$ 2,619 (en)& 11 \\
     SkillMatch-1K~\cite{skillmatch1k} & single & 900 (en) $\times$ 2,648 (en) & 1 \\
    \addlinespace
    \textbf{Candidate Rec.} \\
     Query-Candidate~\cite{jouanneau2024skill} & multi & 200 (en) $\times$ 4,019 (x) & 5 \\
     Project-Candidate~\cite{jouanneau2024skill} & multi & 200 (en) $\times$ 4,019 (x) & 5 \\
    \addlinespace
    \textbf{Job Normalization} \\
     JobBERT~\cite{jobbert} & single & 15,463 (en) $\times$ 2,942 (en) & 24 \\
     MELO (48 datasets)~\cite{retyk2024melo_mels} & multi & 4,438 (bg) $\times$ 150,140 (x) & 21 \\
    \addlinespace
    \textbf{Skill Normalization} \\
     ESCO Alternatives~\cite{esco,workbench} & multi & 72,008 (en) $\times$ 13,939 (en)& 28 \\
     MELS (8 datasets)~\cite{retyk2024melo_mels} & multi & 4,381 (sv) $\times$ 100,273 (en) & 5 \\
    \addlinespace
    \textbf{Skill Extraction} \\
     House~\cite{skillextraction_decorte2022design} & multi & 262 (en) $\times$ 13,891 (en)& 28 \\
     Tech~\cite{skillextraction_decorte2022design} & multi & 338 (en) $\times$ 13,891 (en)& 28 \\
     SkillSkape~\cite{skillskape} & multi & 1,191 (en) $\times$ 13,891 (en)& 28 \\
    \bottomrule
  \end{tabular}%
  }
\end{table}

\subsection{Models \& Baselines}

The WorkRB toolkit provides inference implementations of diverse models and baselines
for recommendation and NLP tasks, 
spanning both neural and traditional retrieval paradigms. On the semantic side, the benchmark provides a \code{BiEncoder} wrapper for any sentence-transformers model~\cite{sentencebert}, along with domain-specialized models contributed from published research, including JobBERT-v1 to v3~\cite{jobbert, contextmatch_jobbertv2, jobbert_v3} and CurriculumMatch~\cite{curriculummatch}, and token-level embedding models such as ConTeXTMatch~\cite{contextmatch_jobbertv2}. 
For lexical baselines, WorkRB includes BM25~\cite{bm25}, TF-IDF, and Edit Distance as traditional retrieval reference points, complemented by a random ranking baseline that establishes a lower-bound reference. 
All models support adaptive target spaces, enabling consistent evaluation across diverse tasks.
While the current release focuses on one-stage embedding-based models, the framework architecture is designed to support broader model families, including multi-stage and generative approaches. 
Table~\ref{tab:results} reports mean average precision (MAP) scores across all task groups for a representative subset of baselines.

\begin{table}[t]
\centering
\caption{MAP per task group. \emph{Multilingual} models are evaluated on all languages with no language aggregation; \emph{EN} models are evaluated on English monolingual subsets only. Best result for both scenarios in \textbf{bold}. 
}
\label{tab:results}
\resizebox{\columnwidth}{!}{%
\begin{tabular}{lccccccc|c}
\toprule
\textbf{Model}                             & \textbf{SRec} & \textbf{ORec} & \textbf{SIRec} & \textbf{CRec} & \textbf{JNorm} & \textbf{SNorm} & \textbf{SExtr}                     & \textbf{All}  \\ \midrule
\textit{\#Multilingual datasets}           & \textit{28}   & \textit{28}   & \textit{12}    & \textit{12}   & \textit{72}    & \textit{35}    & \multicolumn{1}{c|}{\textit{84}}   & \textit{271}  \\ \addlinespace
Random Ranking                             & 0.4           & 0.6           & 0.7            & 8.2           & 0.3            & 0.1            & \multicolumn{1}{c|}{0.1}           & 1.5           \\
BM25~\cite{bm25}                           & 2.9           & 6.1           & 11.0           & 27.5          & 2.5            & 47.4           & \multicolumn{1}{c|}{5.5}           & 14.7          \\
Qwen3-Embed{ \footnotesize (0.6B)}~\cite{qwen3embedding}           & 6.0           & 14.7          & 26.3           & \textbf{50.4} & 12.5           & \textbf{65.5}  & \multicolumn{1}{c|}{\textbf{16.9}} & 27.5          \\
JobBERT-v3~\cite{jobbert_v3}               & \textbf{9.3}  & \textbf{20.5} & \textbf{30.6}  & 43.4          & \textbf{25.6}  & 62.5           & \multicolumn{1}{c|}{16.6}          & \textbf{29.8} \\ \midrule
\textit{\#EN-only datasets}                    & \textit{1}    & \textit{1}    & \textit{2}     & \textit{2}    & \textit{2}     & \textit{1}     & \multicolumn{1}{c|}{\textit{3}}    & \textit{12}   \\ \addlinespace
ConTeXTMatch~\cite{contextmatch_jobbertv2} & 14.6          & 32.2          & 31.8           & 54.5          & 36.7           & 86.8           & \multicolumn{1}{c|}{46.7}          & 43.3          \\
JobBERT-v2~\cite{contextmatch_jobbertv2}   & 15.4          & 33.2          & \textbf{37.4}  & \textbf{56.8} & \textbf{38.9}  & 83.2           & \multicolumn{1}{c|}{27.5}          & 41.8          \\
CurriculumMatch~\cite{curriculummatch}     & \textbf{15.8} & \textbf{34.1} & 34.4           & 52.1          & 36.7           & \textbf{88.2}  & \textbf{47.3}                      & \textbf{44.1} \\
\bottomrule
\end{tabular}
}
\end{table}

\subsection{Multilingual Support}

Beyond multiple tasks, work-domain applications typically require support over multiple languages.
Scaling evaluation over the quadratic task-language evaluation matrix is challenging
as query and target spaces may be defined in different languages, ontologies carry distinct multilingual structures, and evaluation must account for both monolingual and cross-lingual settings.
WorkRB addresses this by dynamically loading ontology structures per language and version, resolving query and target spaces independently. This allows users to define arbitrary monolingual setups (e.g., German queries against German targets) or cross-lingual setups (e.g., French queries against English targets) within the same task. For example, the ESCO ontology~\cite{esco} provides occupation and skill vocabularies in up to 28~languages, all of which WorkRB can load on demand for any ESCO-based tasks.
Additionally, this design allows support for inherent cross-lingual tasks such as the candidate recommendation tasks operating across five languages simultaneously~\cite{jouanneau2024skill,jouanneau2025efficient}, while MELO and MELS~\cite{retyk2024melo_mels} provide cross-lingual normalization datasets that map national taxonomy entries to ESCO across 21 and 5 languages, respectively.
Users can further define monolingual and cross-lingual metric aggregation strategies, as discussed in Section~\ref{sec:design_usage}.

\section{Design \& Usage}
\label{sec:design_usage}

\par{\noindent\textbf{Usage.}} WorkRB follows a simple three-step workflow: (1)~initialize a model, (2)~select tasks and languages, and (3)~run evaluation. The framework is installable via ``\texttt{pip install workrb}'';  Listing~\ref{lst:usage} illustrates the core workflow.

\begin{sloppypar} %
\par{\noindent\textbf{Extensibility.}} A registry system ensures that all tasks and models are dynamically discoverable, where organizations can locally extend WorkRB with proprietary datasets and models, without the need to share sensitive data.
Additionally, WorkRB is designed for sustained collaborative development: structured contribution guidelines, issue-based task proposals, and continuous integration facilitate community-driven growth of both tasks and models.
Extensibility is achieved through abstract base classes. Custom tasks inherit from \code{RankingTask}, implement \code{load_dataset()}, and register via \code{@register_task()}. Tasks map languages to dataset identifiers through \code{languages_to_dataset_ids()}, supporting multiple datasets per language for configurations with same-language datasets (e.g. across regions), and cross-lingual evaluation. 
Similarly, custom models inherit from \code{ModelInterface}, implement \code{compute_rankings()}, and register via \code{@register_model()}.
\end{sloppypar}

\begin{lstlisting}[caption={WorkRB usage example for model evaluation.},label={lst:usage},float=t]
from workrb import models, tasks, evaluate

model = models.BiEncoderModel("all-MiniLM-L6-v2")
tasks = [
 tasks.ESCOSkillNormRanking(split="val", languages=["en"]),
 tasks.MELORanking(split="val", languages=["de", "fr"]),
]
results = evaluate(model, tasks)
\end{lstlisting}

\par{\noindent\textbf{Checkpointing.}} WorkRB provides automatic checkpointing, saving results after each task--dataset completion. Checkpoints track \code{(task, dataset_id)} tuples,  so re-running with the same output folder seamlessly resumes from where evaluation left off. 

\par{\noindent\textbf{Metric aggregation }} follows a four-level hierarchy: (1)~macro-averaging across languages per task, with configurable aggregation modes for monolingual-only or cross-lingual grouping by input or output language; (2)~macro-averaging across tasks per task group (e.g., all skill extraction tasks); (3)~macro-averaging across task groups per task type (e.g., all ranking tasks); and (4)~macro-averaging across task types for the overall benchmark score (e.g. ranking and classification). 
Structured output files include \code{results.json} for metrics, \code{checkpoint.json} for completion state, and \code{config.yaml} for configuration. The default ranking metrics are MAP, NDCG, R-Precision@10, and MRR, with additional metrics available such as Recall@K and Hit@K.
\looseness=-1

\section{Broader Impact}

\subsection{A Community-Driven Ecosystem}

As illustrated in Figure~\ref{fig:teaser}, WorkRB is sustained by three complementary pillars: industry partners contribute models and datasets for real-world tasks, academic labs contribute methodological and foundational innovations, and government institutions maintain the multilingual occupational ontologies (e.g., ESCO, O*NET) that form the taxonomic backbone of many tasks.
This ecosystem is designed around a mutual incentive model: contributors gain visibility through citations and adoption of their resources, while in return they benefit from (i) scalable evaluation across tasks and languages with automated checkpointing, (ii) a standardized interface, (iii) discovery of state-of-the-art models and datasets, and (iv) the ability to extend the benchmark with proprietary tasks for internal use without disclosing sensitive data.
To sustain growth, WorkRB provides clear contribution guidelines, issue-based task proposals, and continuous integration, with a community benchmark challenge planned at 2026 RecSys in HR workshop (pending acceptance).
\looseness=-1

\subsection{Standardization towards Compliant \& Representative Evaluation}

\par{\noindent\textbf{Regulatory landscape.}} Employment data constitutes personal data subject to increasingly strict regulation~\cite{10.1145/3696457}. In the EU, the GDPR constrains processing, while the AI Act classifies employment AI as high-risk, mandating data governance, accuracy, transparency, and conformity assessments, with ongoing standardisation efforts targeting these areas~\cite{jrc_standardisation}. In the US, comparable obligations arise from the CCPA, state-level laws such as New York's Local Law~144~\cite{nyc_ll144}, and EEOC algorithmic hiring guidelines.

\par{\noindent\textbf{Open standardization.}} Open-source benchmarks are well-positioned to support compliance in this landscape.
They enable independent verification, reproducible auditing, and transparent standardization, properties that proprietary benchmarks cannot offer. For this reason, the European Commission explicitly encourages the development of benchmarks and measurement methodologies in the context of accuracy and robustness requirements for high-risk AI systems. WorkRB embodies this principle by building its public tasks on openly licensed data, while its extensible design allows organizations to evaluate proprietary datasets and models internally under the same framework. 
Note, however, that legal compliance extends beyond benchmark accuracy. In the EU, AI providers must also address risks to fundamental rights such as dignity, autonomy, and non-discrimination. Such requirements are difficult to capture in computational metrics, making interdisciplinary assessment essential in practice.
Public Employment Services (PES) show the practical relevance of this discussion. Through its 74 members across five regions, the World Association of PES (WAPES) provides a global platform for exchange on how employment services respond to labor market change, including AI. In this context, transparent, responsible, and context-sensitive evaluation approaches can support peer learning and promote the adoption of trustworthy AI in PES.

\par{\noindent\textbf{Multilinguality fosters representativeness.}} Work-domain AI is deployed worldwide, yet evaluation resources remain concentrated in high-resource languages. By centralizing multilingual evaluation across up to 28~languages through multilingual ontologies and diverse contributed datasets, WorkRB improves accessibility for underrepresented language populations, and shifts focus from English-only evaluations.

\section{Conclusion \& Future Work}
We presented WorkRB, the first community-driven, open-source evaluation framework for AI in the work domain, addressing the fragmentation of evaluation in this multi-stakeholder domain. WorkRB unifies 13~recommendation and NLP tasks with dynamic multilingual ontology support, an extensible pip-installable toolkit, and a multi-stakeholder ecosystem bridging academia, industry, and public institutions.
Future work includes extending task coverage to occupational activities, additional ontologies, and temporal evaluation dimensions, as well as incorporating career path recommendation
and multi-branch architectures including large language models and re-ranking approaches.
As the framework is designed to scale through continued community participation, we encourage contributions for new tasks, datasets, models, and metrics, as exemplified by the many contributions in this paper shaped by multiple stakeholders. Contributing is simple through available getting-started instructions and contribution guidelines at \url{https://github.com/techwolf-ai/WorkRB}.



\end{document}